\documentclass{article}
\usepackage{spconf,amsmath,graphicx}
\usepackage{algorithmic,algorithm}
\usepackage[marginal]{footmisc}

\title{ILLUMINATION ADAPTIVE PERSON REID BASED ON TEACHER-STUDENT MODEL AND ADVERSARIAL TRAINING}
\name{Ziyue Zhang$^1$ \quad Richard Yi Da Xu$^{1\star}$ \quad Shuai Jiang$^1$\quad Yang Li$^1$ \quad Congzhentao Huang$^1$\quad Chen Deng$^1$}
\address{$^1$University of Technology Sydney}
\begin{document}
%
\maketitle
\footnote{\noindent © 2020 IEEE.  Personal use of this material is permitted. Permission from IEEE must be obtained for all other uses, in any current or future media, including reprinting/republishing this material for advertising or promotional purposes, creating new collective works, for resale or redistribution to servers or lists, or reuse of any copyrighted component of this work in other works.}

\begin{abstract}
Most existing works in Person Re-identification (ReID) focus on settings where illumination either is kept the same or has very little fluctuation. However, the changes in the illumination degree may affect the robustness of a ReID algorithm significantly. To address this problem, we proposed a Two-Stream Network that can separate ReID features from lighting features to enhance ReID performance. Its innovations are threefold: (1) A discriminative entropy loss to ensure the ReID features contain no lighting information. (2) A ReID Teacher model trained by images under “neutral” lighting conditions to guide ReID classification. (3) An illumination Teacher model trained by the differences between the illumination-adjusted and original images to guide illumination classification. We construct two augmented datasets by synthetically changing a set of predefined lighting conditions in two of the most popular ReID benchmarks: Market1501 and DukeMTMC-ReID. Experiments demonstrate that our algorithm outperforms other state-of-the-art works and particularly potent in handling images under extremely low light.
\end{abstract}


\section{Introduction}
\label{sec:intro}
Person ReID is a critical computer vision task that aims to match the same person in images or video sequences.
Most researchers in this area focus on the problem of how to deal with people occlusion \cite{huang2018adversarially,hou2019vrstc}, different poses \cite{zhu2019progressive,qian2018pose}, views \cite{zheng2017unlabeled,zheng2019joint} and resolutions. Few have studied illumination variation in spite of its importance. Bhuiyan et al. \cite{bhuiyan2015exploiting} proposed Cumulative Weighted Brightness Transfer Functions to model illumination and viewpoint variations. Wang et al. \cite{wang2014camera} proposed a feature projection matrix which can be learned to project the person images of one camera to the features space of the other camera, so as to conquer device differences in the practical surveillance camera network. Zeng et al. \cite{zeng2019illumination} proposed an illumination-Identity Disentanglement network to separate different scales of illuminations apart. 

Illumination changes in person ReID should be much larger than what the database entails. A person may appear under a specific lighting condition and then reappear under a drastically different lighting condition. Hence, it is crucial to study robust algorithms that can handle changes in illumination degrees.
Under these conditions, both the probe and gallery images are thought to be under some random effects of illumination changes. 

Current person ReID datasets are usually collected under different cameras in a short time frame, such as Market1501 \cite{zheng2015scalable} and DukeMTMC-ReID \cite{ristani2016MTMC,zheng2017unlabeled}, both with very few illumination changes. Because there are no public datasets containing images under different lighting conditions, we construct two augmented datasets from the above two using Gamma correction \cite{4379008} and Poisson noise. By altering the Gamma, each image can produce multiple images under different illumination conditions. We refer them as Market1501-illu and DukeMTMC-illu.

There is one major challenge in the problem we try to solve: when lighting conditions vary across the images, its effect can greatly affect the ReID results as the lighting information entangles with the actual information needed for ReID, making the model less robust against illumination changes as a result. To this end, inspired by \cite{liu2018exploring}, we design a unique network, which separates the illumination features and ReID features. These two features are then fed into the ReID and illumination encoders respectively. To make the model work effectively, we have added two additional innovations summarised below:

\begin{figure*}[tbp]
    \centering
    \includegraphics[width = 1\textwidth]{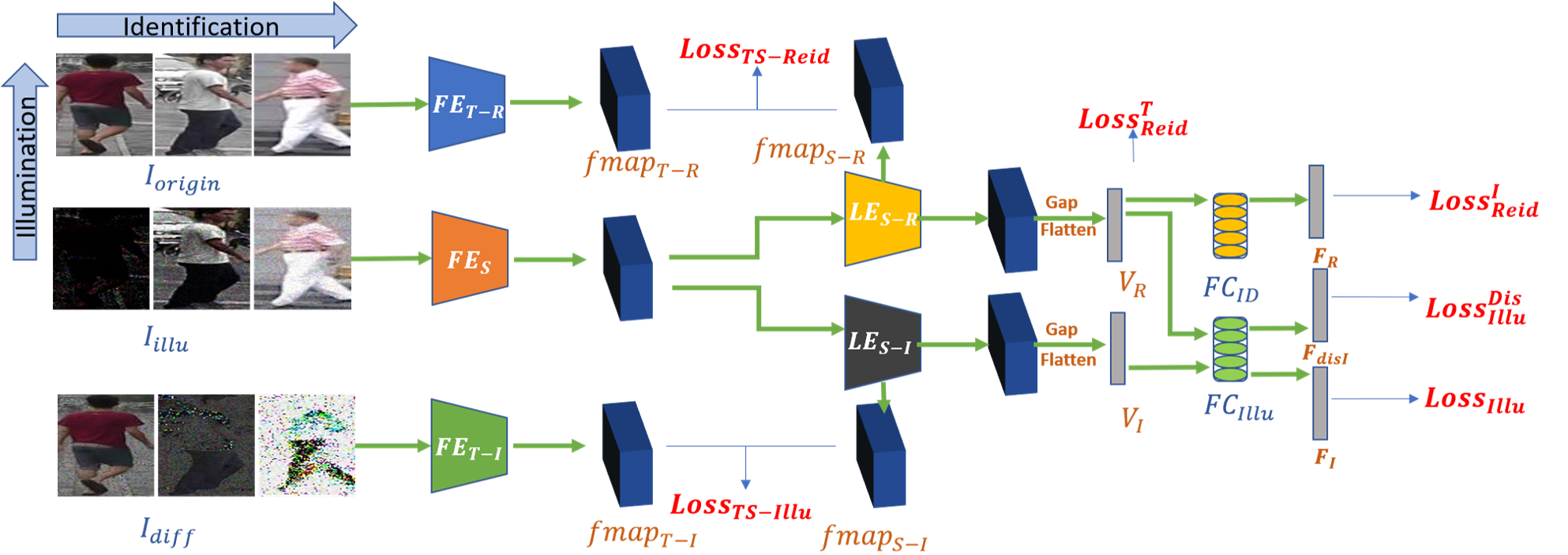}
    \caption{The architecture overview of our TS-D model. It contains (1) Backbone ${FE}_{S}$, ${LE}_{S-R}$ and ${LE}_{S-I}$, (2) ReID Teacher model ${FE}_{T-R}$, (3) Illumination Teacher model ${FE}_{T-I}$ and (4) Illumination variation adaptive module.}
    \label{fig:wholeModels}
\end{figure*}

(1) In order to ensure the ReID features are entirely disentangled from the lighting information, a discriminative high entropy loss is employed. Specifically, we need the ReID features to be classified not only accurately in the ReID classifier, but also "vaguely" from the illumination classifier. This is to make sure there is no residual illumination information remained after disentanglement.

(2) The Disentanglement Network itself may be limited by its ability to separate the two sources of information. Therefore, inspired by \cite{chen2017person, zheng2019joint}, we added two separate Teacher models to help the network improve its disentanglement capabilities:
the ReID Teacher model is trained under "neutral" lighting conditions for varying IDs beforehand. It is then used to guide its Student counterpart in training.
Likewise, to better single-out the illumination, the illumination Teacher model is trained using the differences between the illumination-adjusted and original images as inputs.


\section{Method}\label{sec:Method}


Fig \ref{fig:wholeModels} shows the overall model structure of our TS-D (Teacher-Student Discriminative) model. It includes four main parts: (1) Backbone, (2) ReID Teacher model, (3) Illumination Teacher model and (4) Illumination variation adaptive module.


\subsection{Backbone}
The backbone takes illumination-varying images as inputs. It consists of two parts. The first part is the former share-weights encoder ${FE}_{S}$, which is the first convolution layer followed by the first two blocks of Resnet50. The second part includes the latter ReID encoder and illumination encoder $LE_{S-R}$ and $LE_{S-I}$ that are both the last two blocks of Resnet50. Then we adopt a Global average pooling (Gap) layer on each feature map (latter encoders' output) that is flattened to get the person ReID latent vector $V_{R}$ and illumination latent vector $V_{I}$. Finally, we use two independent fully connected layers $FC_{ID}$ and $FC_{Illu}$ to convert each vector to person ReID factor $F_{R}$ and illumination factor $F_{I}$. 

To classify a person's identity, we adopt two loss functions, which are soft-label identity classification loss $\operatorname{Loss}_{\text{Reid}}^{I}$ for $F_{R}$ and batch hard triplet loss $\operatorname{Loss}_{\text{Reid}}^{T}$ \cite{schroff2015facenet, hermans2017defense} for $V_{R}$.

The first loss function is defined as
\begin{equation}
\operatorname{Loss}_{\text{Reid}}^{I}(F_{R}, N_{person})=  -\sum_{y_{reid}=1}^{N_{person}} p_i^{reid} \log q_i^{reid},
\end{equation}
where
\begin{equation}
    p_{i}^{reid}=\left\{\begin{array}{ll}{1-\frac{N_{person}-1}{N_{person}} \varepsilon} & {\text { if } i=y_{reid}} \\ {\varepsilon / N_{person}} & {\text { otherwise, }}\end{array}\right.
\end{equation}
and $q_i^{reid}$ is the ID prediction logits of current class $i$, $y_{reid}$ is the truth ID label, $N_{person}$ is person identities' number and $\varepsilon$ is a tiny weights factor. By allocating $\varepsilon / N_{person}$ on those false label classes in the training dataset, we can prevent the model from over-fitting the training dataset .

The second loss function is defined as
\begin{equation}
\operatorname{Loss}_{Reid}^{T}(V_{R}) = \sum_{\alpha, p, n \atop y_{a}=y_{p} \neq y_{n}}\left[m+D_{a, p}-D_{a, n}\right]_{+},
\end{equation}
where $D_{a, p}$ and $D_{a, n}$ are the Euclidean distances between the anchor and the positive/negative sample and $m$ is a margin parameter. 

Hence we have the overall person ReID loss function as
\begin{equation}
\operatorname{Loss}_{\text{Person}}=\lambda_{1} \operatorname{Loss}_{\text{Reid}}^{I}+\lambda_{2} \operatorname{Loss}_{\text{Reid}}^{T},
\end{equation}
where $\lambda_{1}$ and $\lambda_{2}$ are adjustable weights.

To classify the illumination condition of each input image, we adopt the illumination classification loss for the illumination factor $F_{I}$, which is denoted by $\operatorname{Loss}_{\text{Illu}}(F_{I}, N_{illu})$ and defined in the same fashion as $\operatorname{Loss}_{\text{Reid}}^{I}(F_{R}, N_{person})$. 



\subsection{Teacher-Student model for person ReID}


Due to the difficulty for the encoder to disentangle image features associated with varying light conditions, we use a ReID Teacher model to guide the learning of the middle block features in $LE_{S-R}$. The ReID Teacher model $FE_{T-R}$ is comprised of the first three blocks of a Resnet50 pretrained on the original dataset and its weights are fixed during training. A regularization loss is constructed to minimize the difference between the outputs of the ReID Teacher model $fmap_{T-R}$ and the middle block outputs of the ReID latter encoder $fmap_{S-R}$:
\begin{equation}
\operatorname{Loss}_{TS-Reid}=\left\|fmap_{\text {T-R}}-fmap_{\text {S-R}}\right\|_{2}^{2}.
\end{equation}

Because the Teacher model is pretrained on original images, it can obtain very high accuracy on the original dataset. Since the input of the Teacher and Student model are identical apart from the illumination condition used, therefore, the Student model’s ability to extract information on altered images can be significantly enhanced by mimicking how the job was done by its Teacher on the middle block features.

\subsection{Teacher-Student model for illumination classification}

To further improve the disentanglement, we additionally use an illumination Teacher model $FE_{T-I}$ to guide the learning of the middle block features in latter illumination encoder $LE_{S-I}$. It has the same structure as the ReID Teacher model. However, it is pretrained on the difference $I_{diff}$ between the original and altered images. Similarly, a regularization loss is constructed between the outputs of the illumination Teacher model $fmap_{T-I}$ and the latter illumination encoder's middle block features $fmap_{S-I}$:
\begin{equation}
\operatorname{Loss}_{TS-Illu}=\left\|fmap_{\text {T-I}}-fmap_{\text {S-I}}\right\|_{2}^{2}.
\end{equation}

By doing this, we can make the middle block features in $LE_{S-I}$ closer to the "natural" lighting features and thus obtain a better illumination classifier. It then uses the adversarial training process to ensure no residual illumination information remained in the ReID feature.

\subsection{Illumination variation adaptive module}

To obtain the illumination-independent features in the ReID stream, we design a discriminative illumination classification loss with the adversarial training process.
At the ReID training stage, we send person ReID vector $V_{R}$ into the weight-fixed illumination classifier ${FC}_{Illu}$. 
We designed an illumination dis-classification loss:
\begin{equation} 
\begin{split}
\operatorname{Loss}_\text{Illu}^\text{Dis}(F_{disI}, N_{illu})= -\sum_{y_{disillu}=1}^{N_{illu}} p_i^{disillu} \log q_i^{disillu} ,
\end{split}
\end{equation}
in which
\begin{equation}
p_{i}^{disillu}=\frac{1}{N}
\end{equation} and $q_i^{disillu}$ is the illumination prediction log-its of current class of $i$ for $F_{disI}$.

Through this loss function, we encourage it to have High entropy, i.e., uniform distribution across all illumination conditions. So person ReID features can not be classified as any condition with certainty. In other words, the person ReID features are fooled to make a non-informative judgment over the illumination conditions. In this way, we can extract the person ReID features which exclude the illumination information.

\subsection{Training and testing}

\begin{algorithm}[h]
\caption{Adversarial training process}
\label{alg1}
\hspace*{0.02in} {\bf Hyper parameter:} 
$\alpha, \lambda1, \lambda2, \beta, P, K, T$    \\
\hspace*{0.02in} {\bf Input:} 
$\{{I_{random}, I_{origin}, I_{diff}}\}$\\
\begin{algorithmic}[1]
\FOR{$i=0, i<={epoch_{max}}, i++$} 
    \STATE For each of K identities, randomly draw P samples as mini-batch
    \IF{i \% T==0}
        \STATE ${loss}_{illu} = {Loss}_{Illu}^{I}+ \alpha * {Loss}_{TS-illu}$  
        \STATE{\textbf{Update} ${loss}_{illu}$}
    \ELSE 
        \STATE Fix $FC_{Illu}$ weights
        \STATE ${loss}_{reid} = \beta * {Loss}_{Illu}^{Dis} +   \lambda1 * {Loss}_{Reid}^{T} + \lambda2 * {Loss}_{Reid}^{I} + \alpha * {Loss}_{TS-Reid}$
        \STATE{\textbf{Update} ${loss}_{reid}$}
    \ENDIF
\ENDFOR
\end{algorithmic}
\end{algorithm}


As shown in Algorithm \ref{alg1}, the whole training process contains two main stages. One is the illumination classification training stage; another is the person ReID training stage. These two training stages will be trained alternatively by our training step $T$.

During testing, we only require the illumination-varying images as the inputs for the ReID backbone. Only the ReID vector $V_R$ is needed for evaluation.

\section{Experiments}\label{sec:Experiment}

\subsection{Experiment setting and Evaluation protocol}
\begin{figure}[htbp]
    \centering
    \includegraphics[width = 0.45\textwidth]{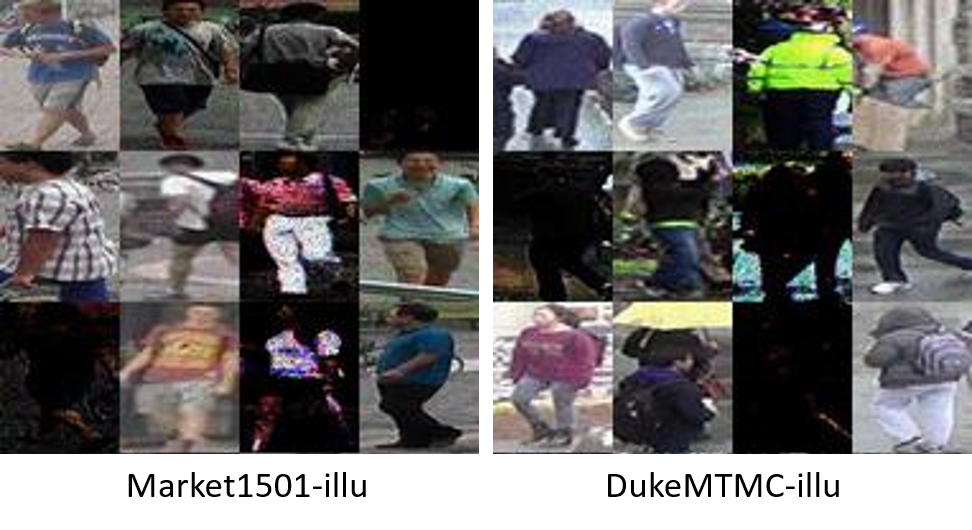}
    \caption{Augmented dataset examples.}
    \label{fig:datasets}
\end{figure}
As mentioned in the Introduction, we generate two illumination ReID datasets Market1501-illu and DukeMTMC-illu based on their original data. For each image, we generate $8$ synthetic images under different levels of illumination by using Gamma correction \cite{4379008} combined with random Poisson noise.
The following gamma values are used: 0.1,0.2,0.3,0.4,0.5,0.7,1,1.5. 
Fig \ref{fig:datasets} shows the augmented datasets. 
Two metrics, the standard Cumulative Matching Characteristics (CMC) values and mean Average Precision (mAP) are adopted for evaluation. 

\subsection{Implementation details}
We use one TESLA V100-SXM2-32GB graphics card for our entire experiments. The batch size is set to 64. We use Adam optimizer with a weight decay of 0.0005. The total training epoch number is set to 240. The base learning rate is set to 0.00035, with a linear scheduler that lowers the rate by using a decay factor 0.1 at epoch 40, 80 and 150. We also use warming up strategy (the first 20 epochs). The weight parameters $\lambda1$ and $\lambda2$ are both set to 1.0. The margin value $m$ in triplet loss is set to 0.3. The weight factor $\alpha$ in the Teacher-Student loss is set to 0.003 and $\beta$ in the discriminative illumination classification loss is set to 0.1. We adopt reranking \cite{zhong2017re} at the testing stage as well.


\subsection{Comparison with state-of-the-art methods}
We exploit the Market1501-illu and DukeMTMC-illu to evaluate the accuracy of our model compared with other current state-of-the-art person ReID deep learning methods including the IID model \cite{zeng2019illumination} that tried solving the illumination problem as well. To the best of our abilities, we re-implemented the IID model using the same baseline as ours to ensure fair comparison. Experimental results are presented in Table \ref{table1}.

\begin{table}[!h]
\caption{Comparison with state-of-the-art ReID methods on Market1501-illu and DukeMTMC-illu.}
\label{table1}
\vspace*{2mm}
\centering
\begin{tabular}{|l|l|l|l|l|l|l|l|}
\hline
model  & \multicolumn{3}{l|}{Market1501-illu} & \multicolumn{3}{l|}{DukeMTMC-illu} \\ \cline{2-7} 
 & R1 & R10 & mAP & R1 & R10 & mAP \\ \hline
MLFN \cite{chang2018multi}  & 68.4 & 86.2 & 44.8 & 65.2 & 81.7 & 45.3 \\ \hline
PCB \cite{sun2018beyond} & 68.1  & 85.7 & 45.4 & 65.4  & 82.9 & 45.1 \\ \hline
HACNN \cite{li2018harmonious} & 51.6  & 78.7 & 29.6 & 46.4  & 71.2 & 29.0 \\ \hline
OSNet \cite{zhou2019osnet} & 73.1 & 87.1 & 48.1 & 68.4  & 82.9 & 43.3 \\ \hline
IID \cite{zeng2019illumination} & 70.7  & 85.7 & 46.7 & 66.9 & 83.7 & 43.6 \\ \hline
TS-D & \textbf{75.5}  & \textbf{88.0} & \textbf{50.7} & \textbf{71.4} & \textbf{85.5} & \textbf{48.9} \\ \hline
\end{tabular}
\end{table}

From the table, we can see that the results on both datasets demonstrate our model achieve the best performance over all other methods.

\subsection{Ablation study}

Our baseline model is a Resnet50 \cite{Luo_2019_CVPR_Workshops}. To illustrate the effectiveness of the proposed modules in our work, we verify five variant settings. TS-D(backbone) is only the two stream output backbone model. TS-D(TS) is the backbone model with two Teacher-Student regularization loss. TS-D(DIS) is the backbone model with discriminative illumination classification loss. TS-D(DIS-TS) is the backbone model with all loss functions. TS-D(DIS-TS-rerank) is the backbone model with all loss functions and reranking in the test stage. The results of the baseline and these variants on Market1501-illu and DukeMTMC-illu are showed in Table \ref{table2}.

\begin{table}[h]
\caption{Performance of baseline and variants of proposed model on Market1501-illu and DukeMTMC-illu.}
\label{table2}
\vspace*{2mm}
\centering
\resizebox{0.5\textwidth}{!}{%
\begin{tabular}{|l|l|l|l|l|l|l|}
\hline
model  & \multicolumn{3}{l|}{Market1501-illu} & \multicolumn{3}{l|}{DukeMTMC-illu} \\ \cline{2-7} 
 & R1 & R10 & mAP & R1 &  R10 & mAP \\ \hline
Baseline & 69.7  & 85.8 & 44.7 & 67.0  & 82.8 & 45.0 \\ \hline
TS-D(backbone) & 71.1  & 86.0 & 47.1 & 68.0  & 83.7 & 45.5 \\ \hline
TS-D(TS) & 72.8  & 86.0 & 49.3 & 69.0  & 83.9 & 46.3 \\ \hline
TS-D(DIS)  & 71.6  & 86.6 & 49.1 & 68.2  & 84.5 & 45.8 \\ \hline
TS-D(DIS-TS)  & 74.7  & 87.9 & 49.8 &  71.3  & \textbf{85.5} & 48.4 \\ \hline
TS-D(DIS-TS-rerank) & \textbf{75.5}  & \textbf{88.0} & \textbf{50.7} & \textbf{71.4}& \textbf{85.5} & \textbf{48.9} \\ \hline
\end{tabular}%
}
\end{table}
Firstly, we can see that the TS-D(backbone) achieves better performance than the baseline on all metrics. This is because the latter independent encoder can separate the illumination and person ID information, which greatly improves the quality of the features sent for person ReID.

Second, TS-D(TS) and TS-D(DIS) improves the backbone to different extents separately. The former demonstrates that these two Teachers can help increase the robustness against illumination fluctuation of the backbone (especially on images under extremely low light) and enhance the ability of illumination classification. The latter shows that the discriminative illumination classification loss can help the ReID latter encoder output ReID features with no illumination information. Combining them together, i.e. TS-D(DIS-TS) obtains better results than using each of them alone, which means these two types of loss can work collaboratively.

Finally, the reranking trick can further improve our complete model a bit, which proves the effectiveness of reranking again as in many existing studies. The total improvements of the proposed model over baseline on all three metrics are 5.8, 2.2, 6.0 on Market1501-illu and 4.4, 2.7, 3.9 on DukeMTMC-illu.

\section{Conclusion}\label{sec:Conclusion}

In this paper, we explored person ReID under different illumination conditions including extremely dark lighting images. Due to the lack of real image datasets with different illumination conditions, we generated two augmented datasets based on Market1501 and DukeMTMC-ReID. Base on the finding that traditional person ReID methods performed inadequately on these augmented datasets, we proposed a novel TS-D ReID model which uses two Teachers and the innovative discriminative loss as well as the adversarial training process to learn the illumination-independent person ReID features. The experimental results demonstrated that our model outperformed other state-of-the-art methods. 


\bibliographystyle{IEEEbib}
\bibliography{refs}

\end{document}